\documentclass[letterpaper, 10 pt, conference]{ieeeconf}  % Comment this line out if you need a4paper
\IEEEoverridecommandlockouts                              % This command is only needed if 
\usepackage{amsmath,graphicx}
\usepackage[utf8]{inputenc} % allow utf-8 input
\usepackage[T1]{fontenc}    % use 8-bit T1 fonts
\usepackage{hyperref}       % hyperlinks
\usepackage{url}            % simple URL typesetting
\usepackage{booktabs}       % professional-quality tables
\usepackage{amsfonts}       % blackboard math symbols
\usepackage{nicefrac}       % compact symbols for 1/2, etc.
\usepackage{microtype}      % microtypography
\usepackage{graphicx}
\usepackage{booktabs}   
\usepackage{xcolor}

\newcommand{\R}{\mathbb{R}}
\newcommand{\Z}{\mathbb{Z}}

\renewcommand{\vec}[1]{\boldsymbol{#1}} 
\newcommand{\mtrx}[1]{\mathbf{#1}}

\title{\LARGE \bf Learning Decision Ensemble using a Graph Neural Network for Comorbidity Aware Chest Radiograph Screening}

\author{Arunava Chakravarty$^{1}$,  Tandra Sarkar$^{2}$, Nirmalya Ghosh$^{1}$, Ramanathan Sethuraman$^{3}$,  Debdoot Sheet$^{1}$% <-this % stops a space
\thanks{*This work is supported through a research grant from Intel India Grand Challenge 2016 for Project MIRIAD.}% <-this % stops a space
\thanks{$^{1}$ A. Chakravarty, N. Ghosh, D. Sheet are with the Indian Institute of Technology Kharagpur, India-721302 ({\tt \{arunava, nirmalya, debdoot\}@ee.iitkgp.ac.in})}%
\thanks{$^{2}$ T. Sarkar is with Apollo Gleneagles Hospital, Kolkata, India}%
\thanks{$^{3}$ R. Sethuraman is with Intel Technology India Pvt. Ltd. Bangalore, India }}

\begin{document}

\maketitle
\thispagestyle{empty}
\pagestyle{empty}

%%%%%%%%%%%%%%%%%%%%%%%%%%%%%%%%%%%%%%%%%%%%%%%%%%%%%%%%%%%%%%%%%%%%
\begin{abstract}
Chest radiographs are primarily employed for the screening of cardio, thoracic and pulmonary conditions. Machine learning based automated solutions are being developed to reduce the burden of routine screening on Radiologists, allowing them to focus on critical cases. While recent efforts demonstrate the use of ensemble of deep convolutional neural networks (CNN), they do not take disease comorbidity into consideration, thus lowering their screening performance. To address this issue, we propose a Graph Neural Network (GNN) based solution to  obtain ensemble predictions which models the dependencies between different diseases. A comprehensive evaluation of the proposed method demonstrated its potential by improving the performance over standard ensembling technique across a wide range of ensemble constructions. The best performance was achieved using the GNN ensemble of DenseNet121 with an average AUC of 0.821 across thirteen disease comorbidities.
\end{abstract}


%%%%%%%%%%%%%%%%%%%%%%%%%%%%%%%%%%%%%%%%%%%%%%%%%%%%%%%%%%%%%%%%%%%%%%%%%%%%%%%%

\section{Introduction}

Chest X-ray radiography (CXR) is a fast and inexpensive imaging modality which is commonly employed for the screening and diagnosis of cardio, thoracic and pulmonary pathologies. The shortage of Radiologists leads to unnecessary delays in the detection of diseases ~\cite{rimmer2017radiologist}, which regresses early intervention. The routine nature of screening is inspiring the development of automated methods in order to prioritize the clinicians time and effort  to the critical cases, as well as reduce the intra- and inter-observer variations in reporting. 

The availability of large public datasets have led to the exploration of different Convolutional Neural Networks (CNNs)~\cite{wang} for multi-label disease classification in CXR images. ResNet-50~\cite{resnet} architecture was adapted in~\cite{baltruschat2019comparison} and additional non-image data (viz. age, gender and the image view) were integrated to improve the classification. An attention guided CNN was explored in~\cite{guan}  where the disease specific regions of interest was estimated first to restrict the classification network's inference to these regions only. Averaging the predictions from an ensemble of multiple CNN models has shown improved performance over single CNNs in~\cite{chexpert} that employed an ensemble of 30 DenseNet~\cite{densenet} models, and in ~\cite{putha2018can} that employed an ensemble of ResNet~\cite{resnet} with squeeze and excitation blocks by varying the model initialization conditions and the training dataset distribution.

Interestingly, chest diseases are pathologically correlated and this observation of joint or otherwise antagonistic appearance of a group of diseases is termed as comorbidity. The presence of a disease class statistically increases/decreases the probability of occurence of other co-related classes (see Fig. \ref{fig:co_occur}). The existing methods have ignored these dependencies with an exception of an unpublished work \cite{yao2017learning} which employed a  Recurrent Neural Network (RNN). However, this method requires the disease classes to be in a fixed order and only models the dependencies of a class with those preceding it. The construction of an ensemble of CNNs that combines the predictions by leveraging the comorbidity dependencies between the different diseases has not been explored so far.

% Contribution

\begin{figure}[]
\centering
	\includegraphics[width=0.48 \textwidth]{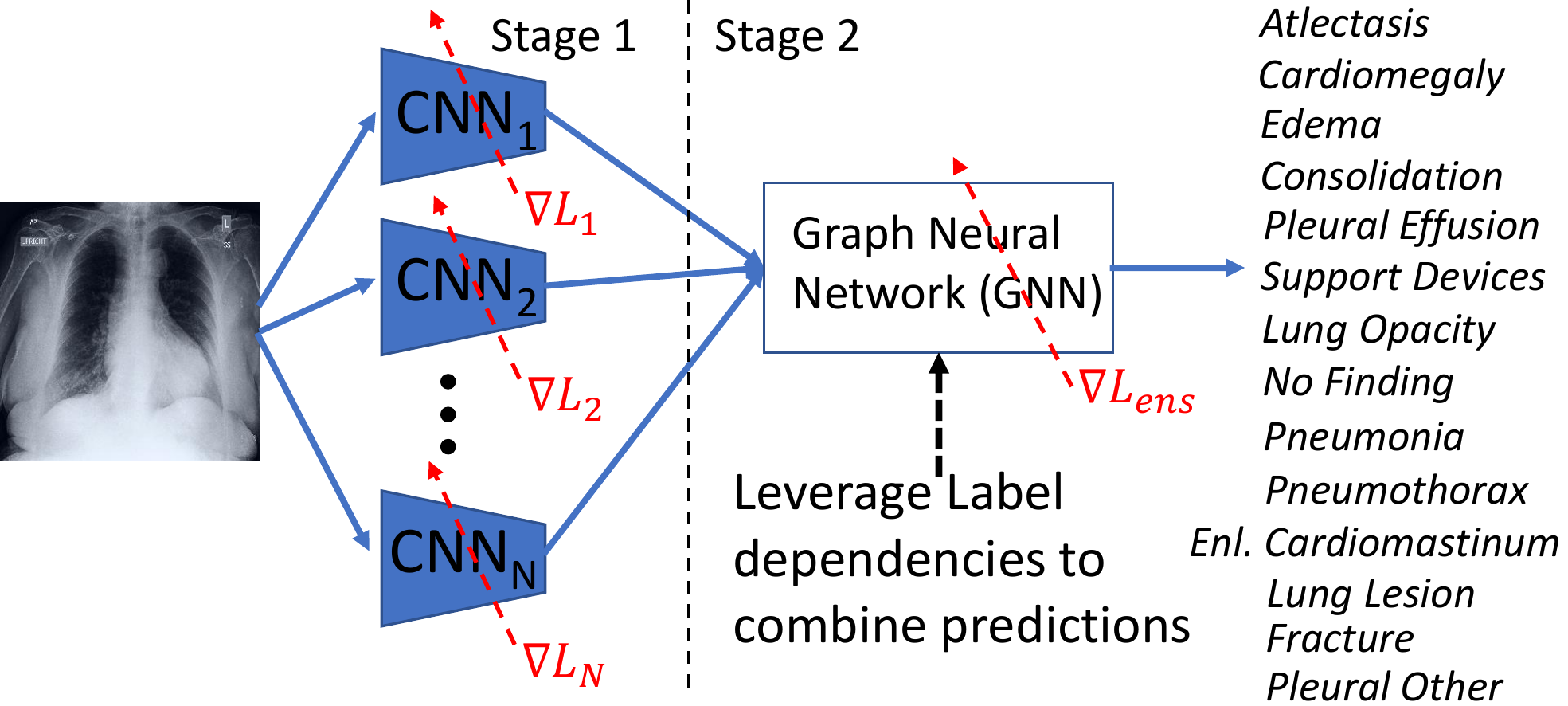}
	\caption{ Decision fusion of an ensemble of CNN models (Stage 1) using a comorbidity prior aware GNN (Stage 2).}
	\label{fig:graphical_abstract}
\end{figure}

In this work (Fig.~\ref{fig:graphical_abstract}), we explore a novel formulation using Graph Neural Networks (GNN) \cite{graph_sage, edge_gnn}  to combine the predictions of an ensemble of CNN models by leveraging the comorbidity statistics. The problem is modeled as a directed weighted graph where each disease class is represented by a vertex and the edge weights define the the degree of co-occurence between each pair of vertices. A comprehensive evaluation of the method is performed by considering ensembles of different CNN architectures constructed by learning multiple network weights for each architecture and using different views of the image.

\section{Method}

\label{sec:method}

\begin{figure*}[h]
\centering
	\includegraphics[width=1 \textwidth]{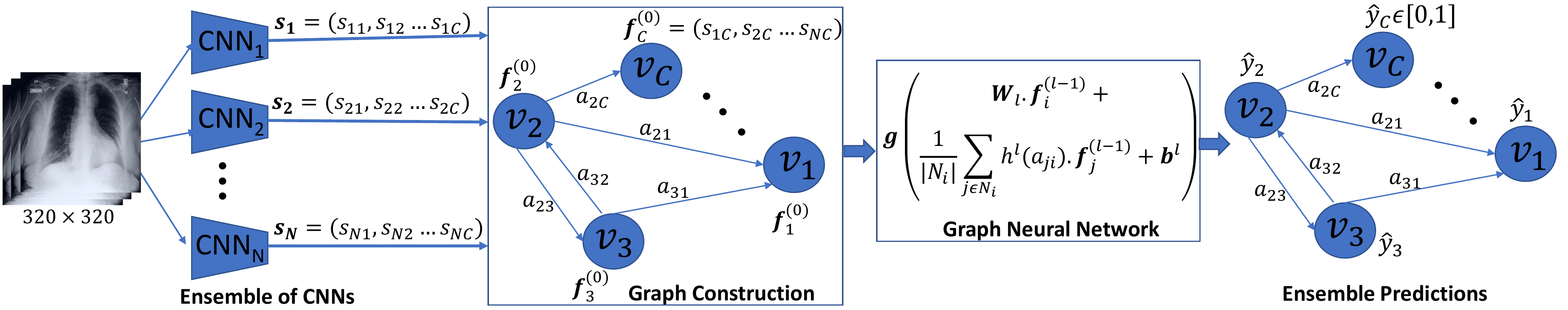}
	\caption{ Overview of the proposed Graph Neural Network Formulation to learn a comorbidity aware decision ensemble.}
	\label{fig:block_diagram}
\end{figure*}

The proposed framework depicted in Fig.\ref{fig:graphical_abstract} is trained in two stages. In stage 1, an ensemble of CNN models is trained to obtain multiple prediction scores (one from each model in the ensemble) for each disease class. The model weights of the CNNs are frozen and the ensemble predictions are combined in stage 2. The task is modeled  using a directed weighted graph to leverage the dependencies between the disease classes. Each disease is represented by a vertex in the graph and a GNN is trained to predict a label for each vertex which denotes the probability of occurrence of the corresponding disease. The details are discussed below.

%\subsection{Ensemble Construction}

\noindent \textbf{Ensemble Construction}: 
%\label{ensemble}
A standard CNN architecture is selected and its final layer is replaced by a fully connected (FC) layer comprising 14 neurons with sigmoid activations. It outputs a \textit{multi-hot} encoding to detect the presence of one or more disease classes. The CNN is initialized with ImageNet~\cite{imagenet} pre-trained weights while FC is randomly initialized. An ensemble for the CNN is constructed by training multiple network weights and employing different veiws of an image during prediction. 
 
A Snapshot Ensembling (SE) \cite{snapshot} approach is employed during training to save multiple network weights for each CNN. The initial learning rate $lr_{mx}$ is decayed to $0$ over a cycle of batch updates using cosine annealing. Next, a warm restart is performed by re-initializing the learning rate to $lr_{mx}$ to allow the network to escape a local minima and the training cycle is repeated multiple times, saving the network weights at the end of each cycle. The weights learned at the end of each training cycle acts as the initialization for the next one.

An ensemble of models is constructed by : i) training a separate CNN on each fold of a four-fold cross-validation on the training set; ii) using SE to obtain multiple network weights and selecting the top $Q$ weights with the highest cross-validation performance for each fold  and iii) employing a 5-crop of the input image (four corner and a central crop) during prediction to obtain a set of 5 predictions for each network weight. Thus, an ensemble of $N=20.Q$ ($4$ folds $\times Q$ weights $\times 5$ crops) network predictions is constructed. 

In Fig. \ref{fig:block_diagram}, the $n^{th}$ prediction in the ensemble denoted by $\vec{s_n}=(s_{n1},s_{n2},...s_{nC}) \in \R ^{C}$ is a \textit{multi-hot} vector, where $C$ is the total number of disease classes and each $s_{ni} \in [0,1]$ is the probability  of the input image to belong to the $i^{th}$ class.

%%%%%%%%%%%%%%%%%%%%%%%%%%%%%%%%%%%%%%%%%%%%%%%%%%%%%%%%
%\subsection{Graph Construction}
\noindent \textbf{Graph Construction}:
%\label{graph_construction}
%We propose to combine the $N$ predictions for an image from the ensemble using a GNN. To this end, 
As depicted in Fig. \ref{fig:block_diagram}, a graph $G(V, A)$ is constructed where $V=\lbrace  v_{i} |  1 \le i \le C, i \in \Z \rbrace$ is a set of $C$ vertices such that the vertex $v_{i}$ corresponds to the $i^{th}$ disease class. An input feature vector $\vec{f_{i}^{(0)}} \in \R ^{N}$ is constructed for each $v_{i}$ by concatenating individual predictions from the $N$ CNNs for the $i^{th}$ class, ie., $\vec{f_{i}^{(0)}}=(s_{1i}, s_{2i}, ... s_{Ni}) \in \R ^{N}$. 

Each element $a_{ij}$ of the adjacency matrix $A \in [-1,1]^{C \times C}$ is the edge weight between $v_{i}$, $v_j$ and is a measure of the degree of co-occurence between the two disease classes measured using the  Cohen's $\kappa$ metric \cite{kappa}. %Mathematically, $a_{ij}=\frac{p_{o}-p{e}}{1-p_{e}}$ where $p_{o}$ is the observed probability of co-occurence and  $p_{e}$ is the probability of chance co-occurence between the disease classes $i$ and $j$. Both $p_o$ and $p_e$ are computed using the Ground Truth (GT) labels of the samples in the training set. 
$\kappa$ is a more robust measure in comparison to percentage agreement as it takes into account the probabilities of the two disease classes to co-occur by random chance. It is bounded in $[-1,1]$ with values close to 1/-1 indicating a strong positive/negative  correlation and 0 indicating independence between the two disease classes \cite{kappa}. As a preprocessing step, the self loops in the graph are removed (by assigning $a_{ii}=0$). Moreover, in order to reduce the computations in the GNN, $A$ is pruned by only retaining the edges to the $K$ neighbors for each node which have the maximum $|a_{ij}|$ values.

%%%%%%%%%%%%%%%%%%%%%%%%%%%%%%%%%%%%%%%%%%%%%%%%%%%%%%%
%\subsection{Graph Neural Network}
\noindent \textbf{Graph Neural Network}: 
%\label{gnn}
The GNN is a deep network with $L$ layers that accepts the graph adjacency matrix $A$ and all vertex features $\vec{f}_{i}^{(0)}$ as input to predict a 1-dimensional label $\hat{y}_{i} \in [0,1]$ for each vertex $v_{i}$. 
Each layer $l \in [1, L]$ performs a Message Passing (MP)  operation on the $d_{l-1}$ dimensional feature representations $\vec{f}^{(l-1)}_{i}$ from the previous $(l-1)^{th}$ layer to compute a $d_{l}$ dimensional feature $\vec{f}^{(l)}_{i}$ for each $v_{i}$. The output of the final $L^{th}$ layer is the 1-dimensional prediction score, ie.,  $\hat{y}_{i}=\vec{f}^{(L)}_{i}$. Mathematically, the MP operation is defined as 
\begin{equation}
\vec{f}_{i}^{(l)}=g\left( \mtrx{W}_{l} . \vec{f}_{i}^{(l-1)}+\frac{1}{|\mathcal{N}_{i}|} \sum _{j \in \mathcal{N}_{i}} h^{l}(a_{ji}). \vec{f}_{j}^{(l-1)}  + \vec{b}_{l} \right),
\end{equation}

where $\mtrx{W} _{l} \in \R ^{d_{l} \times d_{l-1}}$ and $\vec{b}_{l} \in \R ^{d_{l}}$ are learnable weights of the $l^{th}$ layer of the GNN. $\mathcal{N}_{i}$ represents the set of immediate neighbors for $v_{i}$ connected by direct edges.  $g()$ is the activation function where $ReLU()$ is employed in all except the final layer where $Sigmoid()$ activation is used to obtain the class prediction scores and  $h^{l}()$ is a fully connected network. 
The MP for each $v_{i}$ comprises three operations: i) The node feature $\vec{f}_{i}^{(l-1)}$ is transformed into a $d_l$ dimensional vector by matrix multiplication with $W_l$. ii) Next, the features $\vec{f}_{j}^{(l-1)}$ from the immediate neighbors of $v_{i}$ are aggregated into a $d_l$ dimensional feature (details of the Aggregation Function is discussed below). iii) The transformed node and the aggregated neighborhood features are added with the bias $\vec{b}_l$ and the activation function $g()$is applied to obtain $\vec{f}_{i}^{(l)}$.

Since, graphs (unlike images or N-D lattices) donot define a specific ordering among the neighbors, the GNNs employ a permutation invariant Aggregation Function. Traditionally, an average or max operation is employed \cite{graph_sage} which leads to a loss of structural information as it treats each neighbor identically without considering the edge-weights. Hence, inspired from \cite{edge_gnn}, we employ a \textit{weighted} summation operation for aggregation. A weight matrix is learned for each feature $\vec{f}_{j}^{(l-1)}$   using a multi-layer perceptron $h^{l}$ which takes the corresponding edge-weight $a_{ji}$ as input. It comprises two fully connected layers. The first layer has  $\lfloor\frac{d_{l} \times d_{l-1}}{2}\rfloor$ neurons with $ReLU()$ activation followed by the second layer whose output is reshaped to a $(d_{l} \times d_{l-1})$ weight matrix. $Tanh()$ operation is used in the final layer to allow negative values.

\section{Experiments}
\label{sec:experiments}

%%%%%%%%%%%%%%%%%%%%%% COMPLETE AUC TABLE FOR ALL CLASSES#############################################################

\renewcommand{\arraystretch}{1.2}
\begin{table*}[!h]
\centering
\caption{ {\footnotesize Area under the ROC curves (AUC) for the Chest X-ray disease classification. The Average AUC across the thirteen disease classes is reported in the last column. The best performance of each architecture is indicated in bold for each disease. (S) denotes a single model, (E) denotes ensemble by averaging predictions and (GNN) denotes the proposed ensembles combined using GNN.}}
\label{Tab_auc_all}
\resizebox{\textwidth}{!}{
\begin{tabular}{@{}lccccccccccccc|c@{}}
\toprule
                 & Atelectasis & \begin{tabular}[c]{@{}l@{}}Cardio-\\ megaly\end{tabular} & Edema  & \begin{tabular}[c]{@{}l@{}}Consolid\\ -ation\end{tabular} & \begin{tabular}[c]{@{}l@{}}Pleural\\ Effusion\end{tabular} & \begin{tabular}[c]{@{}l@{}}Support\\ Devices\end{tabular} & \begin{tabular}[c]{@{}l@{}}Lung\\ Opacity\end{tabular} & \begin{tabular}[c]{@{}l@{}}Enlarged\\  Cardiom.\end{tabular} & \begin{tabular}[c]{@{}l@{}}No\\ Finding\end{tabular} & \begin{tabular}[c]{@{}l@{}}Pneum-\\ onia\end{tabular} & \begin{tabular}[c]{@{}l@{}}Pneumo-\\ thorax\end{tabular} & \begin{tabular}[c]{@{}l@{}}Lung\\ Lesion\end{tabular} & \begin{tabular}[c]{@{}l@{}}Pleural \\ Other\end{tabular} & Avg.\\ \midrule
                 
ResNet18 (S) \cite{resnet}&  0.721 & 0.735  & \textbf{0.916}  & 0.893 & 0.931 & 0.904  &  0.910 & 0.462 &  0.857 & 0.622  & 0.729  &  0.189 &  0.893 &   0.751\\ 
ResNet18 (E) & 0.756 & 0.787 & 0.909 & \textbf{0.907} &  \textbf{0.938} &  \textbf{0.943} & \textbf{0.927}  & 0.488 & 0.886  & \textbf{0.733} & 0.839  &  0.017 &  0.944 &  0.775 \\
ResNet18 (GNN) & \textbf{0.773} & \textbf{0.821} & 0.906 & 0.870 & 0.937 & 0.936 & 0.926  &  \textbf{0.615} &  \textbf{0.894} & 0.502 & \textbf{0.850}  & \textbf{0.657}  &  \textbf{0.983} &  \textbf{0.820} \\ \midrule

DenseNet121(S) \cite{densenet} & 0.746 & 0.781 & 0.912 & \textbf{0.939} & 0.937 & 0.934 & 0.918  & 0.463 & 0.881  & 0.611 &  0.807 & 0.017  &  0.944 &   0.761\\ 
DenseNet121(E)  & 0.764 & 0.787 & \textbf{0.924} & 0.923 & \textbf{0.944} & \textbf{0.954} &  \textbf{0.932} & 0.523 & \textbf{0.884}   & \textbf{0.677} & 0.835  & 0.069  &  0.953 & 0.782  \\
DenseNet121(GNN)& \textbf{0.785} & \textbf{0.799} & 0.908 & 0.922 &  0.942 &  0.948 & 0.931  & \textbf{0.627} &  0.865 & 0.597 &  \textbf{0.858} & \textbf{0.528}  &  \textbf{0.966} & \textbf{0.821}  \\ \midrule

Xception(S) \cite{xception} & 0.781 & 0.762 & 0.899 & \textbf{0.911} & 0.926 & 0.923 & 0.910 & 0.465 & 0.873 & 0.655 & 0.862 & 0.288 & 0.914 & 0.782  \\
Xception(E) & 0.772 & 0.788 & \textbf{0.916} & 0.907 & \textbf{0.940} & \textbf{0.948} & \textbf{0.928} & 0.475 & 0.878 & \textbf{0.679} & 0.863 & 0.150 & \textbf{0.966} & 0.785 \\ 
Xception(GNN)& \textbf{0.786} & \textbf{0.835} & 0.915 & 0.865 & 0.933 & 0.941 & 0.916 & \textbf{0.586} & \textbf{0.879} & 0.575 & \textbf{0.910}  & \textbf{0.476}  & 0.897  & \textbf{0.810}  \\ 
\bottomrule
\end{tabular}
}
\end{table*}

\begin{figure}[t]
\centering
	\includegraphics[width=0.45 \textwidth]{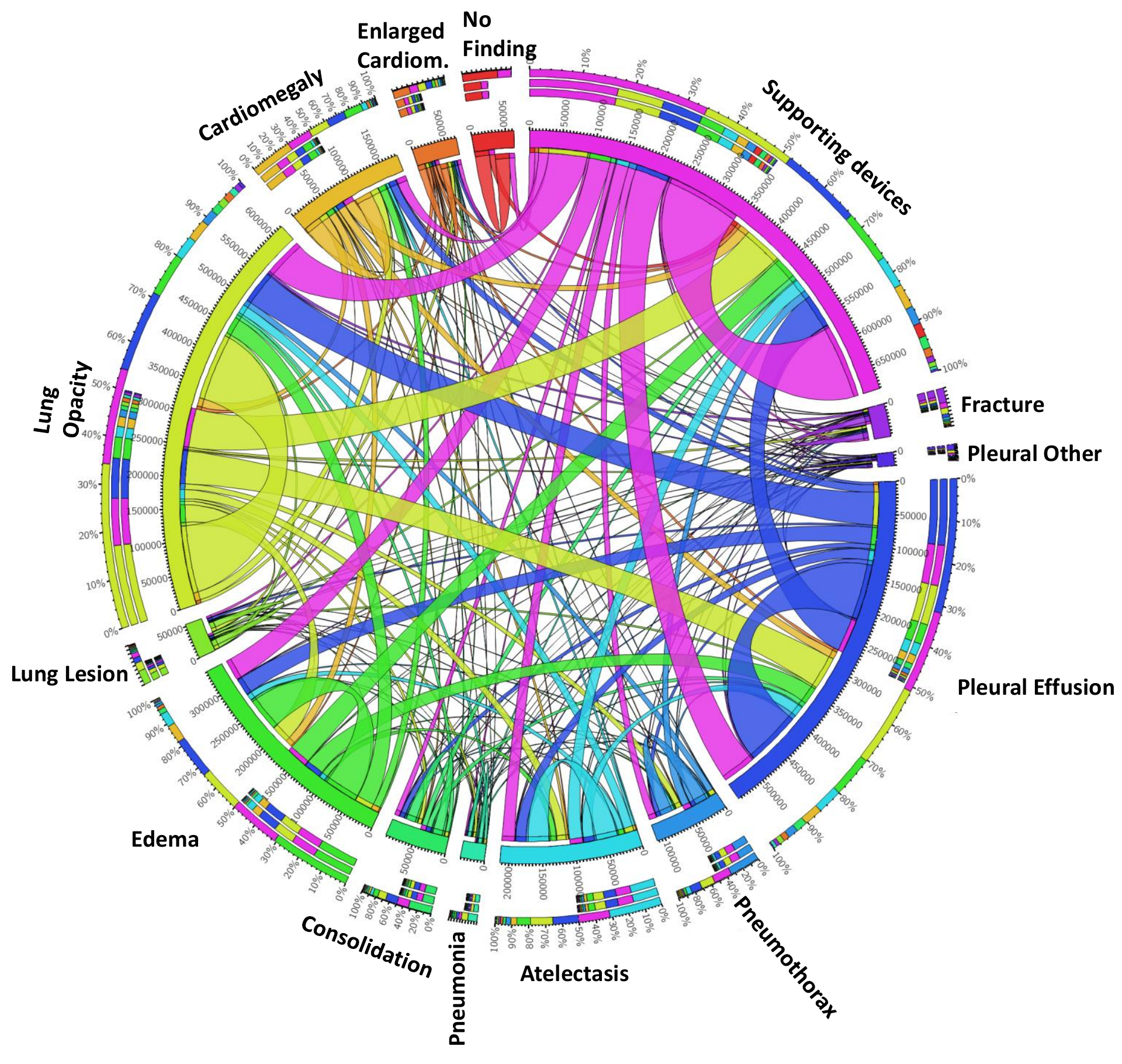}
	\caption{Chord diagram representing the distribution of fourteen comorbid chest diseases in the Chexpert training dataset. The classes occupy an arc length along circumference proportional to its frequency. The thickness of the links connecting a pair of classes indicates their degree of co-occurence.}
	\label{fig:co_occur}
\end{figure}

\textbf{Dataset:} The proposed method has been evaluated on the CheXpert dataset \cite{chexpert} which consists of $223,414$ training and $234$ test images with Ground Truth (GT) labels for 14 diseases. The GT for the training set is noisy and labeled as either present (1), absent (0) or uncertain (-1) as they were automatically  obtained from free-text radiology reports. In our experiments, the uncertain labels were treated as the absence of the disease. The dependencies between the various classes is depicted in Fig. \ref{fig:co_occur}. %similar to the \textit{U-Zeros} setting in \cite{chexpert}
 GT for the test set did not have uncertain labels and obtained from the majority consensus opinion of 3 Radiologists~\cite{chexpert}. There are no samples of the ``Fracture" class in the test set.

\textbf{Training:} The \textit{Binary cross-entropy loss} is used to train each CNN in stage 1 and the GNN in stage 2. The input 2D grayscale chest radiographs are pre-processed by resizing them to  $320 \times 320$ and replicating to obtain a 3-channel input for the CNNs. The channels are normalized to match the statistics of the ImageNet~\cite{imagenet} dataset. Data augmentation comprising  random horizontal flips and  random crops followed by resize operation are applied to the training images on-the-fly. The experiments were run on a server with $2\times$ Intel Xeon 4110 CPU, $12\times 8$ GB DDR4 RAM, $4\times$ Nvidia GTX 1080Ti GPU with $11$ GB RAM and Ubuntu 16.04 operating system. The models were implemented in Python using the Pytorch 1.0 and Pytorch Geometric \footnote{{https\://pytorch-geometric.readthedocs.io/}} library. 

The CNN models in the ensemble were trained using SE for 7 cycles (each cycle is of 2 epochs with 10,647 batch updates per epoch, batch size of 16 and $lr_{mx}=10^{-4}$) using the Adam optimizer~\cite{adam} to select the top $Q=2$ model weights with highest cross-validation performance.  

The GNN was trained for  $8$ epochs, $22,341$ batch updates per epoch with a batch size of $8$ using Adam~\cite{adam} optimizer, learning rate of $10^{-4}$ and a weight decay of $1\times 10^{-5}$.

\textbf{Result:} A comprehensive evaluation of the proposed method is performed on ensembles constructed with identical CNN architecture but different network weights learned using SE on four folds of the training set and using five-crop views for each test image. The ResNet-18 \cite{resnet}, DenseNet-121 \cite{densenet} and the Xception \cite{xception} architectures have been considered.

The hyperparameters for the GNN were empirically fixed through experimentation: (i) $k$ neighbors considered for each vertex was fixed to 5 for ResNet, DenseNet and 9 for the Xception ensembles. (ii) Number of layers $L$ was fixed to 5 for ResNet, 8 for DenseNet and 6 for Xception ensembles. (iii) For all the three ensembles, the input vertex features dimensionality  $d_{0}=40$, the dimensionality $d_{1}$ of the output of $1^{st}$ layer was fixed to 30 and the feature dimensionality was progressively increased across the layers as $d_{l}=\lfloor 1.3 \times d_{l-1} \rfloor, \forall 2 \le l \le L-1$ for all ensembles, with $d_{L}=1$ in the final $L^{th}$ layer to obtain the class predictions.

% To edit from here: <Need to discuss the results in the Table>
The Area under the ROC curve (AUC) for each disease class in the test set is reported in Table 1 \footnote{Due to space limitations, the Sensitivity, Specificity metrics and ROC plots are available online at \url{http://bit.do/Suppl_EMBC_GNN}}. The baseline average ensembles used the same set of CNN models as employed in stage 1 of the proposed method but obtained the final ensemble decision by averaging the predictions of the individual CNN models instead of employing a GNN. Considering the average AUC values across all the thirteen disease classes reported in the last column of Table 1, we make the following observations. Both the GNN and the baseline ensemble models performed superior to the corresponding single model in terms of the average AUC values. Furthermore, \textit{The proposed GNN based ensembles consistently outperformed the corresponding baseline ensembles with an improvement of $\left( \frac{0.820-0.775}{0.775} \times 100= \right) 5.8\%$  for ResNet, $\left( \frac{0.821-0.782}{0.782} \times 100= \right) 4.99\%$ for DenseNet and $\left( \frac{0.810-0.785}{0.785} \times 100= \right) 3.19\%$ for the Xception architecture ensembles}. Among the three GNN ensembles, DenseNet performed the best (AUC=$0.821$) closely followed by ResNet (AUC=$0.820$) while the Xception ensemble had a marginally lower performance (AUC=$0.810$). A qualitative evaluation of the region where the DenseNet ensemble attended for classification was performed by treating the entire ensemble as a black box and employing the Randomized  Input  Sampling  for  Evaluation  (RISE) \cite{rise} to compute the saliency maps. The saliency maps for the GNN based ensembles were in general found to be closer to the manual annotations by a Radiologist in comparison to the average baseline ensemble on a subset of test images (see Fig. 4 for few examples). %Few examples are presented in Fig. xx.

\begin{figure}[h]
 \centering
  \includegraphics[width=.4\textwidth]{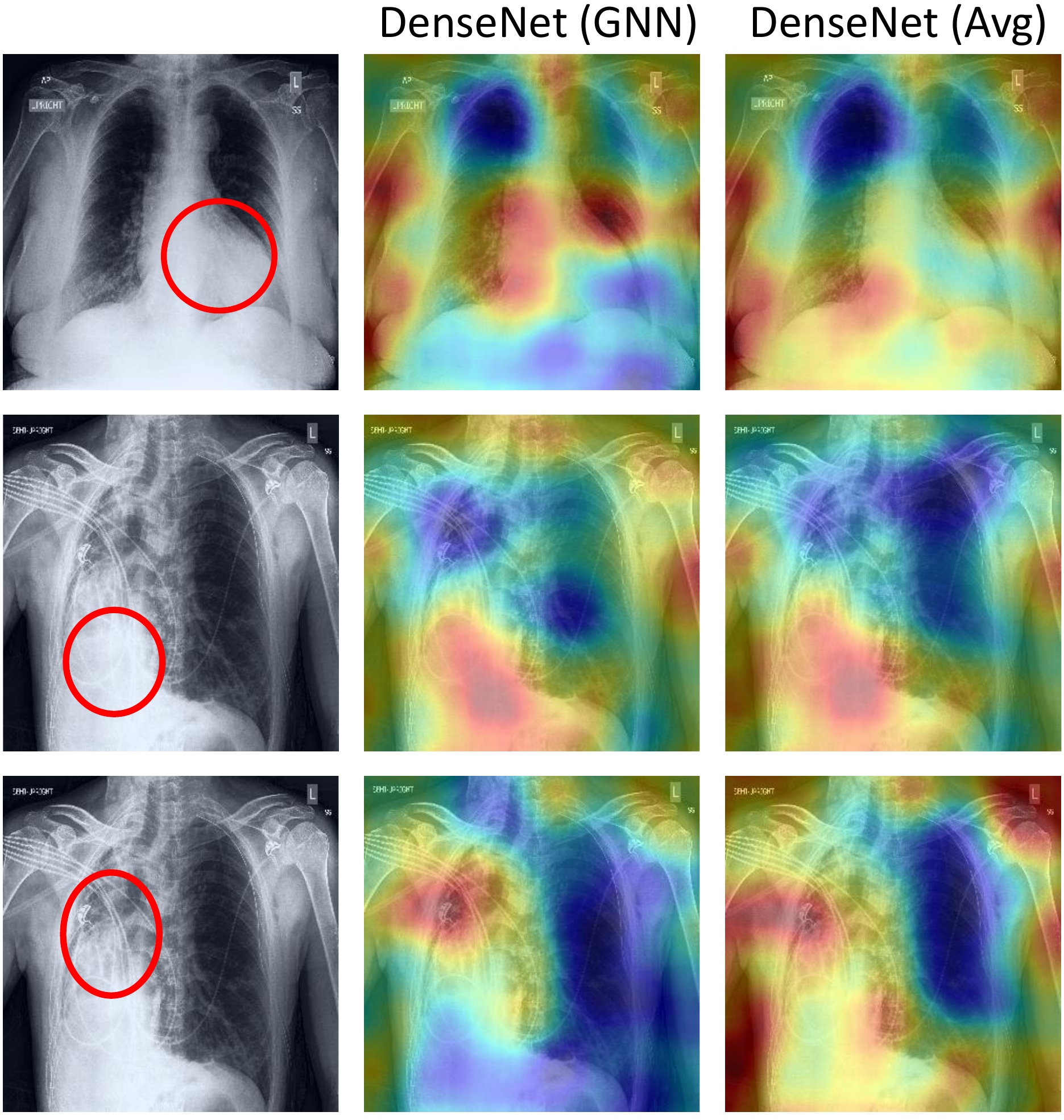}
 \caption{RISE \cite{rise} based saliency map visualization for the proposed GNN (column 2) and baseline average ensembles (column 3) of DenseNet-121. The abnormal region is marked in \textsc{Red} by a Radiologist (column 1). Top row: Cardiomegaly; Middle row: Pleural Other; Bottom row: Lung Opacity. Middle and Bottom rows are comorbid diseases in the same CXR image.}
\label{Fig:RISE} 
 \end{figure}

\vspace{-7pt}
\section{Conclusion}
We explored a novel GNN based framework to obtain ensemble predictions by modeling the dependencies between different diseases in chest radiographs. A comprehensive evaluation of the proposed method demonstrated its potential by improving the performance over standard ensembling technique across a wide range of ensemble constructions. The best performance was achieved using the GNN ensemble of DenseNet121 with an average AUC of 0.821 across thirteen disease comorbidities. A systematic search over the  hyperparameters of the GNN consisting of the number of layers, feature dimensionality in each layer, K nearest neighbors and the number of models used to construct the ensemble may further improve the performance. Since, the graph adjacency matrix was constructed using the noisy labels in the training set that were obtained using automated NLP tools, a clinical validation/correction of these dependencies may be performed in the future. 
%  ################# BIBLIOGRAPHY ###################################
\bibliographystyle{unsrt}
\bibliography{root}

% ########################Supplementary#######################################
%\FloatBarrier 
\newpage
\onecolumn
\setcounter{page}{1}
\section*{Supplementary Materials: Learning Decision Ensemble using a Graph Neural Network for Comorbidity Aware Chest Radiograph Screening}

\begin{figure*}[!ht]
 \centering
  \includegraphics[width=1.0\textwidth]{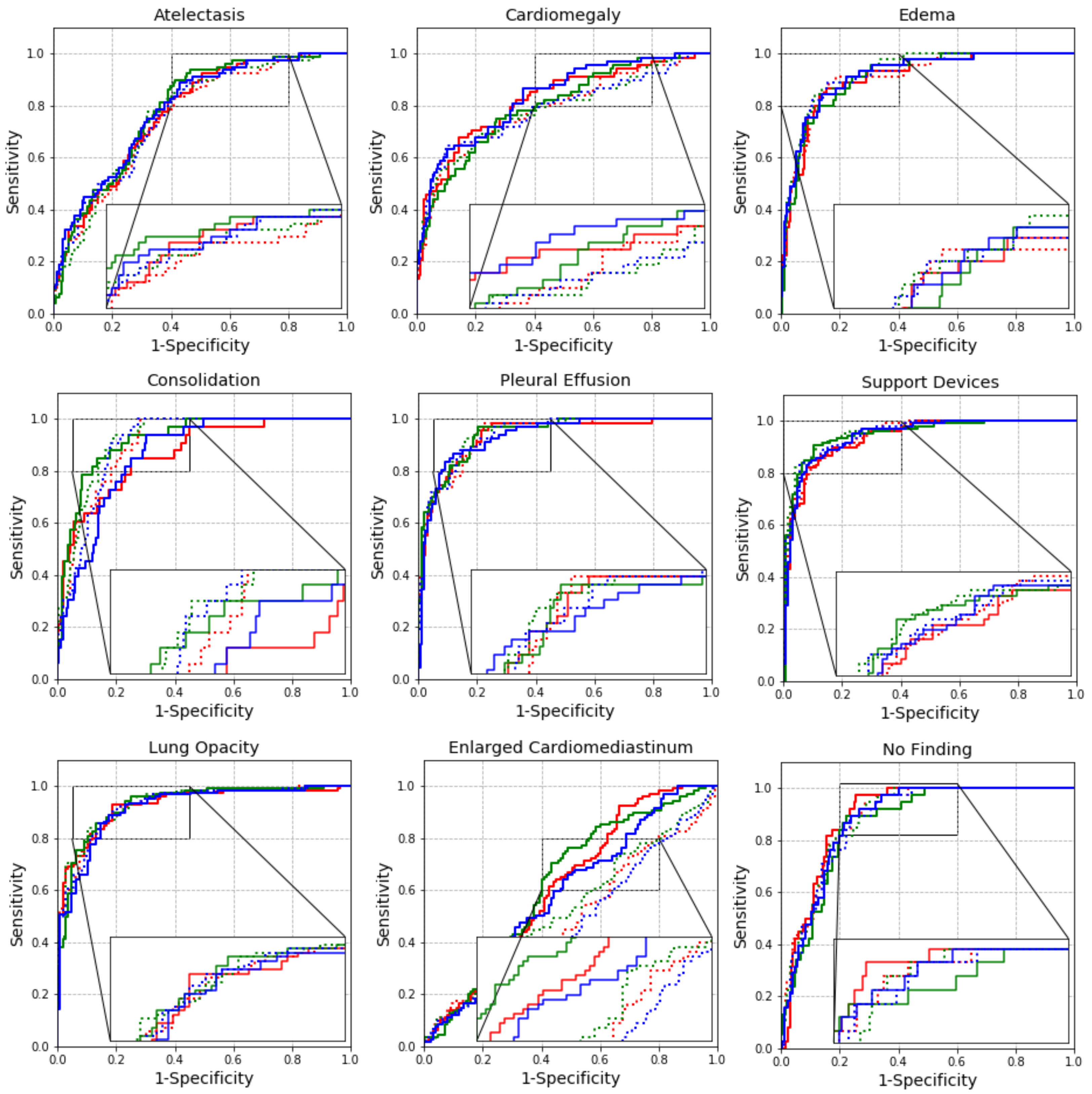}
 \caption{The ROC plots for the nine disease classes. The performance of the proposed GNN ensemble constructed using ResNet is plotted in \textsc{Red}, DenseNet in \textsc{Green} and Xception in \textsc{Blue}. The corresponding baseline average ensembles are plotted using dotted lines of the same color.}
\label{Fig:ROC} 
 \end{figure*}

\renewcommand{\arraystretch}{1.2}
\begin{table*}[!h]
\centering
\caption{Sensitivity (Sens.) and Specificity(Spec. ) for the Chest X-ray disease classification. The operating point on the ROC curve was selected to maximize the the Youden's index (J=Sens.+Spec. -1). (S) denotes a single model, (E) denotes ensemble by averaging predictions and (G) denotes the proposed ensembles combined using GNN.}
\label{Tab__sens_spec_all}
\resizebox{1\textwidth}{!}{
\begin{tabular}{@{}lcccccccccccccccccccccccccccc@{}}
\toprule
         & \multicolumn{2}{c}{Atelectasis} & \multicolumn{2}{c}{Cardiomegaly} & \multicolumn{2}{c}{Edema} & \multicolumn{2}{c}{Consolidation} & \multicolumn{2}{c}{Pleural Effusion} & \multicolumn{2}{c}{Support Devices} & \multicolumn{2}{c}{Lung Opacity} & \multicolumn{2}{c}{Enlarged Cardiom.} & \multicolumn{2}{c}{No Finding} &  \multicolumn{2}{c}{Pneumonia} &  \multicolumn{2}{c}{Pneumothorax} &  \multicolumn{2}{c}{Lung Lesion} &  \multicolumn{2}{c}{Pleural Other} \\ 
         & Sens. & Spec. & Sens. & Spec. & Sens. & Spec. & Sens. & Spec. & Sens. & Spec. & Sens. & Spec. & Sens. & Spec. & Sens. & Spec. & Sens. & Spec. & Sens. & Spec.   & Sens. & Spec.  & Sens. & Spec.  & Sens. & Spec.  \\ \midrule
         
ResNet18(S) & 0.938 & 0.448 & 0.559 & 0.867 & 0.889 & 0.873 & 0.939 & 0.741 & 0.910 & 0.826 & 0.888 & 0.772 & 0.738 & 0.944 & 0.055 & 0.984 & 0.895 & 0.750 & 0.750 & 0.633 & 0.500 & 0.876 & 1.000 & 0.189 & 1.000 & 0.893 \\ 
ResNet18(E)  & 0.825 & 0.591 & 0.588 & 0.916 & 0.911 & 0.815 & 1.000 & 0.716 & 0.985 & 0.778 & 0.850 & 0.898 & 0.881 & 0.833 & 0.174 & 0.896 & 0.947 & 0.714 & 0.750 & 0.743 & 0.875 & 0.673 & 1.000 & 0.017 & 1.000 & 0.944 \\ 
ResNet18(G) & 0.825 & 0.610 & 0.676 & 0.855 & 0.867 & 0.847 & 0.848 & 0.751 & 0.955 & 0.784 & 0.822 & 0.921 & 0.929 & 0.815 & 0.927 & 0.336 & 0.974 & 0.745 & 0.375 & 0.903 & 1.000 & 0.624 & 1.000 & 0.657 & 1.000 & 0.983  \\ \midrule

DenseNet121(S)  & 0.825 & 0.591 & 0.618 & 0.861 & 0.889 & 0.847 & 0.939 & 0.851 & 0.866 & 0.874 & 0.944 & 0.850 & 0.810 & 0.870 & 0.917 & 0.152 & 1.000 & 0.730 & 0.625 & 0.664 & 0.750 & 0.965 & 1.000 & 0.017 & 1.000 & 0.944  \\ 
DenseNet121(E)  & 0.775 & 0.682 & 0.691 & 0.789 & 0.867 & 0.862 & 0.939 & 0.811 & 0.925 & 0.838 & 0.916 & 0.874 & 0.841 & 0.898 & 0.422 & 0.656 & 0.947 & 0.709 & 0.750 & 0.695 & 0.750 & 0.805 & 1.000 & 0.069 & 1.000 & 0.953\\ 
DenseNet121(G) & 0.900 & 0.584 & 0.750 & 0.729 & 0.800 & 0.873 & 0.848 & 0.866 & 0.970 & 0.796 & 0.907 & 0.898 & 0.857 & 0.870 & 0.706 & 0.568 & 0.895 & 0.776 & 0.375 & 0.934 & 1.000 & 0.580 & 1.000 & 0.528 & 1.000 & 0.966  \\ \midrule

Xception(S)   & 0.800 & 0.688 & 0.662 & 0.849 & 0.911 & 0.820 & 1.000 & 0.701 & 0.746 & 0.952 & 0.850 & 0.843 & 0.825 & 0.870 & 0.193 & 0.856 & 0.921 & 0.745 & 0.750 & 0.602 & 0.750 & 0.854 & 1.000 & 0.288 & 1.000 & 0.914  \\ 
Xception(E)  & 0.850 & 0.578 & 0.647 & 0.898 & 0.889 & 0.841 & 0.909 & 0.821 & 0.881 & 0.856 & 0.841 & 0.945 & 0.905 & 0.815 & 0.101 & 0.952 & 0.895 & 0.765 & 0.625 & 0.823 & 1.000 & 0.611 & 1.000 & 0.150 & 1.000 & 0.966 \\ 
Xception(G)  & 0.888 & 0.571 & 0.632 & 0.898 & 0.844 & 0.862 & 0.939 & 0.697 & 0.836 & 0.910 & 0.850 & 0.906 & 0.889 & 0.824 & 0.661 & 0.512 & 0.921 & 0.735 & 0.250 & 0.960 & 1.000 & 0.730 & 1.000 & 0.476 & 1.000 & 0.897 \\
\bottomrule
\end{tabular}
}
\end{table*}

%\fi  %% This comments out the entire table

\end{document}